\documentclass[10pt,twocolumn,letterpaper]{article}

\usepackage{iccv}
\usepackage{times}
\usepackage{epsfig}     %

\usepackage{graphicx}
\usepackage{amsmath}
\usepackage{amssymb}
\usepackage{booktabs}
\usepackage{algorithm}
\usepackage[noend]{algpseudocode}
\algdef{SE}{Begin}{End}{\textbf{begin}}{\textbf{end}}

\DeclareMathOperator*{\argmin}{arg\,min}
\usepackage{booktabs}
\usepackage{array}
\usepackage{multirow}
\usepackage{nicefrac}       %
\usepackage{microtype}  
\usepackage{tabularx}
\usepackage{float}
\usepackage{enumitem}\usepackage{wrapfig,booktabs}
\usepackage{amsfonts}
\usepackage{pifont}%
\usepackage[pagebackref,breaklinks,colorlinks]{hyperref}

\usepackage[capitalize]{cleveref}
\crefname{section}{Sec.}{Secs.}
\Crefname{section}{Section}{Sections}
\Crefname{table}{Table}{Tables}
\crefname{table}{Tab.}{Tabs.}

\newcommand{\LPIPS}{\textbf{LPIPS$\downarrow$}}  %
\newcommand{\MSE}{\textbf{MSE$\downarrow$}}      %
\newcommand{\w}{{\rm\bf w}}         %

\usepackage{afterpage}
\usepackage{scalerel,xparse}
\usepackage{capt-of,etoolbox}
\usepackage{amsmath}
\usepackage[pagebackref,breaklinks,colorlinks]{hyperref}
\usepackage[capitalize]{cleveref}
\crefname{section}{Sec.}{Secs.}
\Crefname{section}{Section}{Sections}
\Crefname{table}{Table}{Tables}
\crefname{table}{Tab.}{Tabs.}
\def\iccvPaperID{8593} %

\iccvfinalcopy
\begin{document}

\title{Make It So: Steering StyleGAN for Any Image Inversion and Editing}
\author{Anand Bhattad\hspace{1cm} Viraj Shah\hspace{1cm} Derek Hoiem\hspace{1cm} D.A. Forsyth\\
University of Illinois Urbana-Champaign\\
\texttt{\href{https://anandbhattad.github.io/makeitso/}{https://anandbhattad.github.io/makeitso/}}
}
\maketitle

\begin{abstract}

StyleGAN's disentangled style representation enables powerful image editing by manipulating the latent variables, but accurately mapping real-world images to their latent variables (GAN inversion) remains a challenge. Existing GAN inversion methods struggle to maintain editing directions and produce realistic results.

To address these limitations, we propose Make It So, a novel GAN inversion method that operates in the $\mathcal{Z}$ (noise) space rather than the typical $\mathcal{W}$ (latent style) space. Make It So preserves editing capabilities, even for out-of-domain images. This is a crucial property that was overlooked in prior methods. 
Our quantitative evaluations demonstrate that Make It So outperforms the state-of-the-art method PTI~\cite{roich2021pivotal} by a factor of five in inversion accuracy and achieves ten times better edit quality for complex indoor scenes.

\end{abstract}

\section{Introduction}
\label{sec:intro}

	\begin{figure}[t!]
		\centering
			\setlength{\linewidth}{\linewidth}
		\setlength{\hsize}{\linewidth}
    \setlength\tabcolsep{0.5pt}
    \renewcommand{\arraystretch}{0.2}
    \begin{tabular}{cccc}
		    \includegraphics[width=0.24\linewidth]{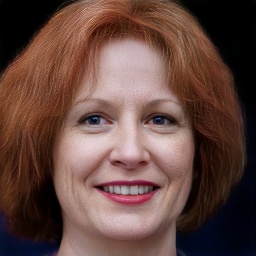} &
		    \includegraphics[width=0.24\linewidth]{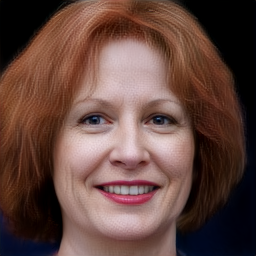}&
		    \includegraphics[width=0.24\linewidth]{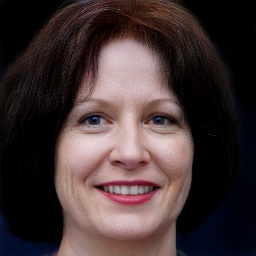} &
		    \includegraphics[width=0.24\linewidth]{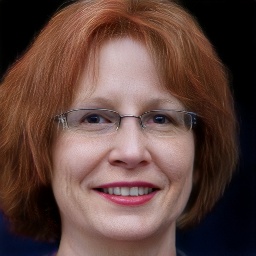}\\

		    \includegraphics[width=0.24\linewidth]{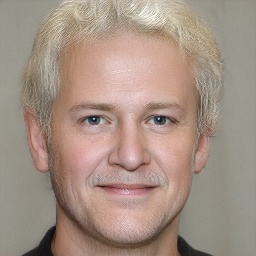} &
		    \includegraphics[width=0.24\linewidth]{figures_supp/face_stylespace3/face1.jpeg}&
		    \includegraphics[width=0.24\linewidth]{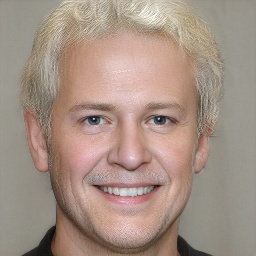} &
		    \includegraphics[width=0.24\linewidth]{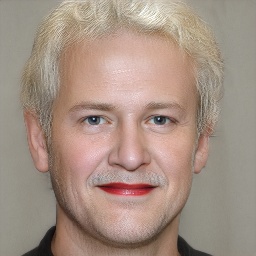}\\
      		    \includegraphics[width=0.24\linewidth]{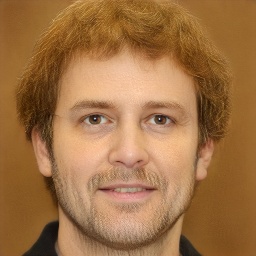} &
		    \includegraphics[width=0.24\linewidth]{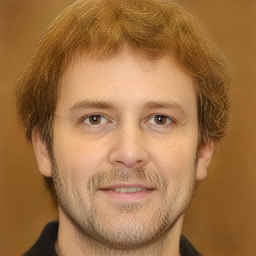}&
		    \includegraphics[width=0.24\linewidth]{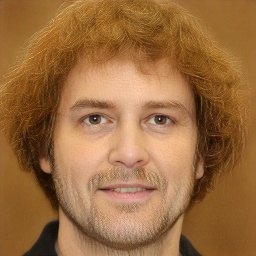} &
		    \includegraphics[width=0.24\linewidth]{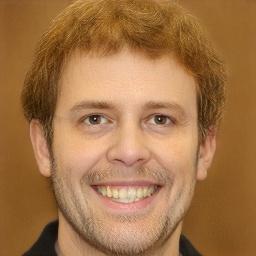}\\
             \includegraphics[width=0.24\linewidth]{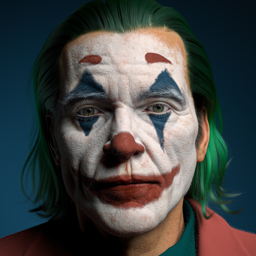} &
		    \includegraphics[width=0.24\linewidth]{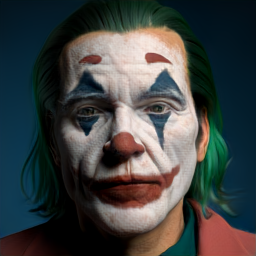}
		    &
		    \includegraphics[width=0.24\linewidth]{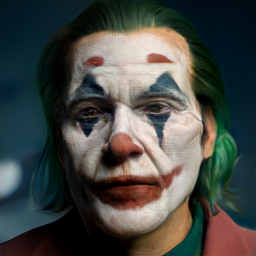}
		    & 
		    \includegraphics[width=0.24\linewidth]{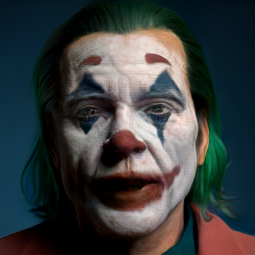} \\

		    \\
            \scriptsize{(1) Target Scene}& 
            \scriptsize{(2) Make It So}&
            \scriptsize{(3) Edited - 1 }&
            \scriptsize{(4) Edited - 2}
        \end{tabular}
        \vspace{5pt}
\caption{Make It So preserves edits in familiar contexts, such as face editing, including hair color, hair style, smile, makeup, and pose, when using a StyleGAN model trained on faces, similar to other GAN inversion methods. These editing directions are obtained from StyleSpace~\cite{wu2021stylespace}. Note each row has two different edits to show a diverse set of editing.}
\label{fig:teaser}
\vspace{-10pt}
\end{figure}

A StyleGAN with parameters $\theta$ maps a standard normal noise vector $z$ to a latent style code $w$ ($w(z)$) and then maps that to an image $x=G(w; \theta)$. The style codes have important semantic properties, making it possible to find directions $s$ that edit the generated image in a semantically meaningful way without losing realism. For example, $G(w+s; \theta)$ might be the same face as $G(w; \theta)$ but now with spectacles. The $s$ that produces a beard, spectacles, or a big nose is independent of the particular $w$ and can be used as a general edit direction. However, these edit directions cannot be applied to real images because we do not know the $z$ that makes the image. The task of GAN inversion is to find the $z$ that corresponds to a given image, allowing us to apply these edit directions to real-world images.

For a GAN inverter to be useful, it needs to possess two crucial properties. First, it must have the accuracy, meaning that $G(w(z_{i}); \theta)$ must be extremely close to $x$ so that we can edit the correct image. Second, it must have edit consistency, meaning that the semantic meaning of $s$ is preserved, so that if $s$ makes a beard, then $G(w(z_i)+s; \theta)$ should show a beard. Without edit consistency, we would have to search for a direction that makes a beard for every new real image, which is impractical. A third property, generalization, is convenient but not essential. Generalization means that we can invert an image in one domain and edit it using a StyleGAN trained on a different domain. This paper introduces the first GAN inverter, Make It So, that possesses all three properties. 

One might invert by seeking a $w_i$ rather than a $z_i$; we find this can increase accuracy but weakens
editability.  Experiments suggest that a given StyleGAN generator cannot produce every in-domain image, which is one reason
really accurate GAN inversion is hard.  Recent practice inverts by finding both a code and a modification to the
StyleGAN, so given $x$, find $z_{i}$, $\theta'$ so that $G(w(z_i); \theta')=x$, resulting in good inversions but weak
edit consistency.  We adopt this approach and show how to manage the search to produce excellent inversions and strong
edit consistency.

\noindent \textbf{Contributions:} We propose Make It So, a novel GAN inversion method that achieves superior accuracy, edit consistency, and generalization for complex scenes compared to existing methods. Our approach inverts images in the noise space ($\mathcal{Z}$ space) using a joint optimization process to find both the best $z$ and the best generative network, unlike previous methods that only fine-tune with a pivot $w$. We introduce anchor and support losses to ensure edit consistency and generalization. Our method is significantly more accurate than the current state-of-the-art method by a factor of five and preserves editing capabilities by a factor of ten. Finally, we demonstrate the ability of our method to produce out-of-domain images, which was not possible with prior methods.

\section{Related Work}
Inverting an image to obtain the corresponding latent code is a critical step in using pre-trained GAN models for image manipulation and editing. Various GAN-based editing methods rely on traversing the latent space to generate meaningful edits, highlighting the importance of the GAN inversion approach for achieving desired manipulations~\cite{zhu2016generative,brock2017neural,Larsen2016AutoencodingBP,perarnau2016invertible,collins2020editing, harkonen2020ganspace, voynov2020unsupervised, shen2020sefa, shen2020interfacegan, patashnik2021styleclip, jahanian2019steerability,Tzelepis_2021_ICCV, Song2022EditingOG}. Several GAN inversion approaches have been proposed~\cite{xia2021gan}. These approaches can be broadly categorized into three categories: optimization-based, encoder-based, and hybrid.

\textbf{Optimization-based methods} find the latent code by minimizing a loss function between the inversion estimate and the target~\cite{Lipton2017PreciseRO, Creswell2019InvertingTG, zhu2016generative,lipton2017precise,karras2020analyzing,abdal2019image2stylegan,abdal2020image2stylegan++}. These methods vary in the loss functions used to measure similarity between the estimate and target, the latent space chosen for the optimization ($\mathcal{Z}$, $\mathcal{W}$, or $\mathcal{W^+}$ space), and additional criteria used to aid optimization (such as improved initialization, stochastic clipping, and early stopping). Although these methods can produce fairly accurate results, their iterative optimization approach is slow and can get stuck in local minima due to the complex and non-convex loss surfaces resulting from large-scale deep generator models.

\textbf{Encoder-based methods} attempt to learn an end-to-end encoder that takes the target image as the input and predicts the latent code directly~\cite{xia2021gan,pidhorskyi2020adversarial,Richardson2020EncodingIS,alaluf2021restyle,Chai2021UsingLS,Larsen2016AutoencodingBP,zhu2016generative,brock2017neural,perarnau2016invertible,tov2021designing,wei2021simpleinversion, wang2021HFGI, xuyao2022, Moon2022IntereStyleEA, Mao2022CycleEO}. Approaches under this category vary in their architecture, loss functions, and training procedures. Encoder-based methods are faster in inference as compared to other methods, but suffer from poor quality results and lack of generalization. Particularly, the encoder-based methods fail heavily even if the domain/alignment of the target image differs ever so slightly from the original - making it useless for editing real scenes.

Hybrid approaches combine both optimization and encoder-based methods \cite{Chai2021EnsemblingWD,Bau2019SeeingWA,zhu2016generative,alaluf2021restyle,bau2019seeing,Bau2019SemanticPM,huh2020transforming, wei2021simpleinversion}. For instance, Parmar et al. \cite{parmar2022spatially} propose a spatially adaptive inversion process that involves multiple layers with segmentation masks. However, these approaches have not investigated the inversion and editing of complex indoor scenes or provide marginal improvements with imperfect inversions.

\noindent \textbf{Inversion and Editing Complex Indoor Scenes:} Indoor scenes, such as bedrooms, pose a significant challenge for GAN inversion due to the presence of multiple unaligned objects. These scenes are more difficult to invert compared to faces because of the presence of a complex spatial relationship between different objects, lighting, and materials. Several inversion approaches have been proposed to this end; Gu et al. \cite{gu2020image} and Kafri et al. \cite{kafri2021stylefusion} attempt to combine features generated by multiple latent codes, while Poirier et al. \cite{poirier2022overparameterization} propose overparameterization of the latent space. Subrtov et al. \cite{Subrtov2022ChunkyGANRI} divide an image into multiple segments, and Xu et al. \cite{xu2021continuity} jointly invert two consecutive images. Kang et al. \cite{kang2021gan} apply geometric transformations to invert out-of-domain scenes, and Kim et al. \cite{kim2021exploiting} and Park et al. \cite{park2020swapping} exploit the generator architecture to achieve better inversion. Bai et al. \cite{Bai2022HighfidelityGI} replace the constant padding in convolution layers with instance-aware coefficients for better inversion.  Despite these efforts, the problem of inversion and editing of complex indoor scenes, including bedrooms, remains open.

	\begin{figure}[t]
		\centering
			\setlength{\linewidth}{\linewidth}
		\setlength{\hsize}{\linewidth}
    \setlength\tabcolsep{0.5pt}
    \renewcommand{\arraystretch}{0.2}
    \begin{tabular}{cccc}
		    \includegraphics[width=0.24\linewidth]{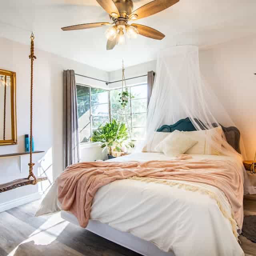} &
		    \includegraphics[width=0.24\linewidth]{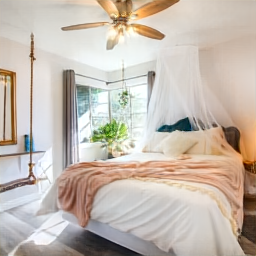}&
		    \includegraphics[width=0.24\linewidth]{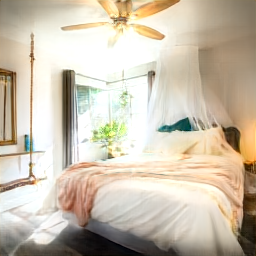} &
		    \includegraphics[width=0.24\linewidth]{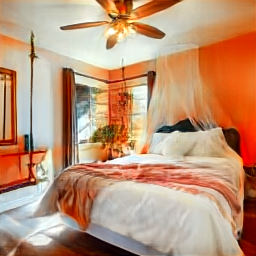}\\

		   	 \includegraphics[width=0.24\linewidth]{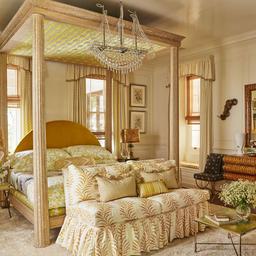} &
		    \includegraphics[width=0.24\linewidth]{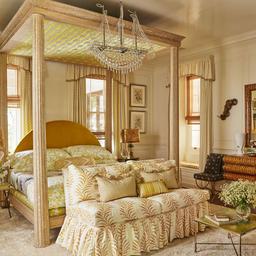}
		    &
		    \includegraphics[width=0.24\linewidth]{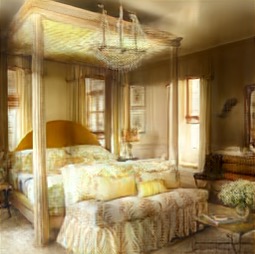}
		    & 
		    \includegraphics[width=0.24\linewidth]{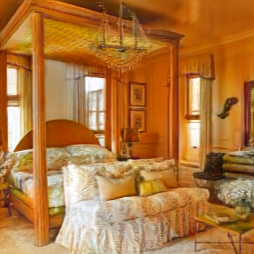} \\
             \includegraphics[width=0.24\linewidth]{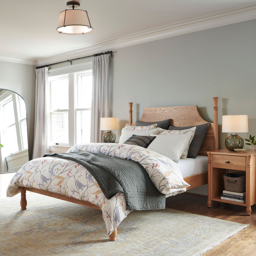} &
		    \includegraphics[width=0.24\linewidth]{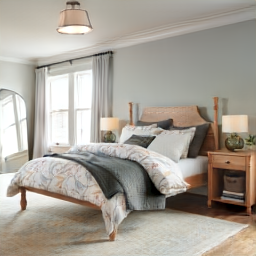}
		    &
		    \includegraphics[width=0.24\linewidth]{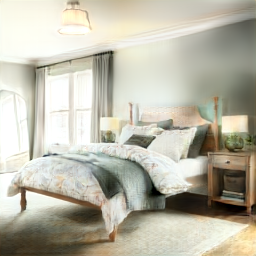}
		    & 
		    \includegraphics[width=0.24\linewidth]{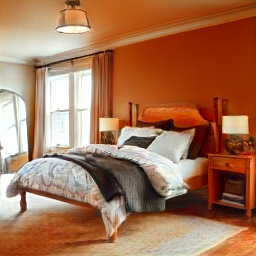} \\

            \includegraphics[width=0.24\linewidth]{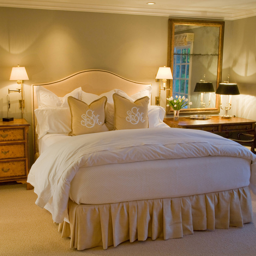} &
		    \includegraphics[width=0.24\linewidth]{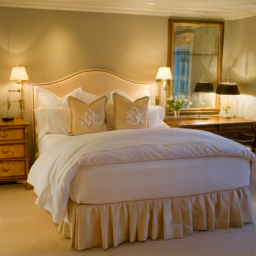}
		    &
		    \includegraphics[width=0.24\linewidth]{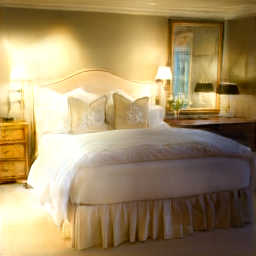}
		    &
		    \includegraphics[width=0.24\linewidth]{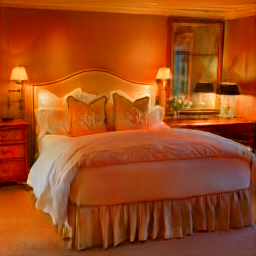}
		    \\
            \scriptsize{(1) Target Scene}& 
            \scriptsize{(2) Make It So}&
            \scriptsize{(3) Relighted}&
            \scriptsize{(4) Resurfacing}
                    \vspace{5pt}
        \end{tabular}

\caption{Make It So further extends to complex indoor scenes. Column 1 displays out-of-distribution bedroom images obtained from the web. Column 2 shows the inversion results obtained by Make It So. Columns 3 and 4 demonstrate the application of edits to the inverted scene after inversion: relighting and resurfacing, respectively. Column 3 highlights the strong lighting effects produced by the approach in all scenes. Column 4 shows realistic surface edits, such as changes in the wall color and hardwood floor surface. These edit directions are obtained from~\cite{bhattad2022enriching}.}
\label{fig:bedroom_teaser}
\vspace{-15pt}
\end{figure}

\begin{figure*}
    \setlength{\linewidth}{\textwidth}
    \setlength{\hsize}{\textwidth}
    \centering
    \includegraphics[width=\textwidth]{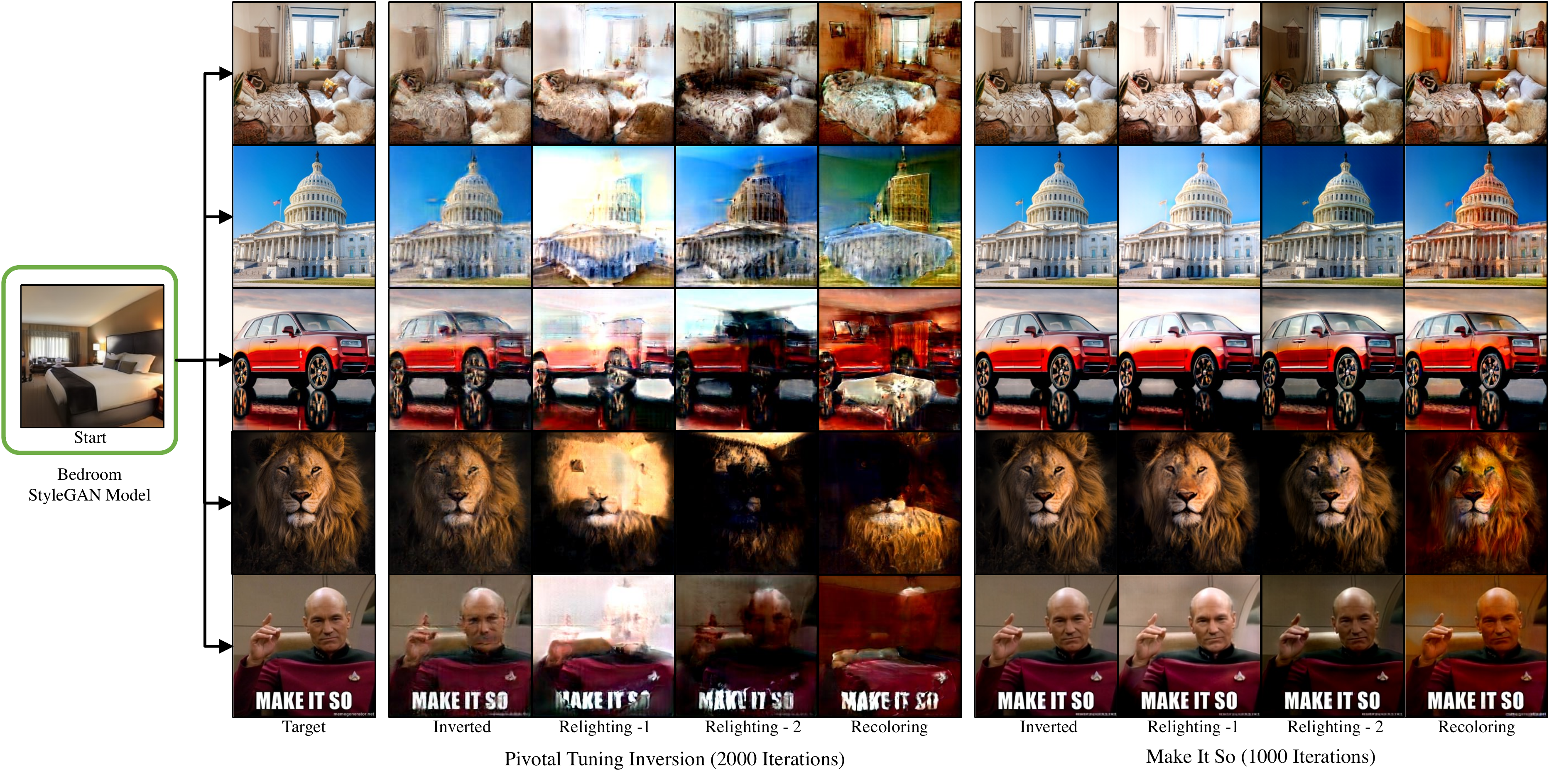}
        \vspace{-15pt}
\caption{We show the generalization capability of Make It So to diverse out-of-domain images beyond complex indoor scenes, using the StyleGAN model trained solely on bedroom images. Our goal is to edit ``Target" images downloaded from the web. Our approach successfully produces realistic image inversion and enables global edits, such as relighting and recoloring, for various out-of-distribution images, including human faces, cars, and indoor and outdoor scenes, using the \emph{same} StyleGAN model trained solely on bedroom images. In comparison, Pivotal Tuning Inversion (PTI~\cite{roich2021pivotal}), a recent optimization and fine-tuning based state-of-the-art, fails to produce realistic edits and produces artifacts such as wall-like patterns in the background and bed-like patterns in the foreground. Our approach outperforms existing methods and achieves high-precision inversion and editing capabilities by inverting in the noise space ($\mathcal {Z}$), rather than the latent style space ($\mathcal{W}$ ) typically used in GAN-based inversion methods. Intuition for why Make It So generalizes well is provided in Fig.~\ref{fig:intuition}.}
\vspace{-10pt}
\label{fig:out-of-domain-generalization}
\end{figure*}

In this work, we aim to address the gap in the inversion and editing of complex indoor scenes with Make It So (see Fig.~\ref{fig:bedroom_teaser}). Editing images with changes to StyleGAN coefficients has a broad reach, as demonstrated by established face editing techniques \cite{shen2020interfacegan, shen2020sefa, wu2020stylespace}, as well as recent work showing that StyleGAN can relight or resurface scenes \cite{bhattad2022enriching}. We use these methods as running examples for editing bedrooms. To relight a scene, we add an offset to the latent style code ($\w^+$) that controls the scene's lighting, producing a realistic image of the scene under different lighting conditions, as demonstrated in column 3 of Fig.~\ref{fig:bedroom_teaser}. Resurfacing or recoloring edits involves making changes to the surface properties of objects in the scene, such as color, texture, and reflectance. This is achieved by adding another offset that controls these properties, resulting in realistic surface edits, such as changes in the wall color and hardwood floor surface, as shown in column 4 of Fig.~\ref{fig:bedroom_teaser}. Furthermore, our method generalizes to out-of-domain images (Fig.~\ref{fig:out-of-domain-generalization}), a previously unseen property in prior inversion methods.

\begin{figure*}[t]
    \centering
    \includegraphics[width=1\textwidth]{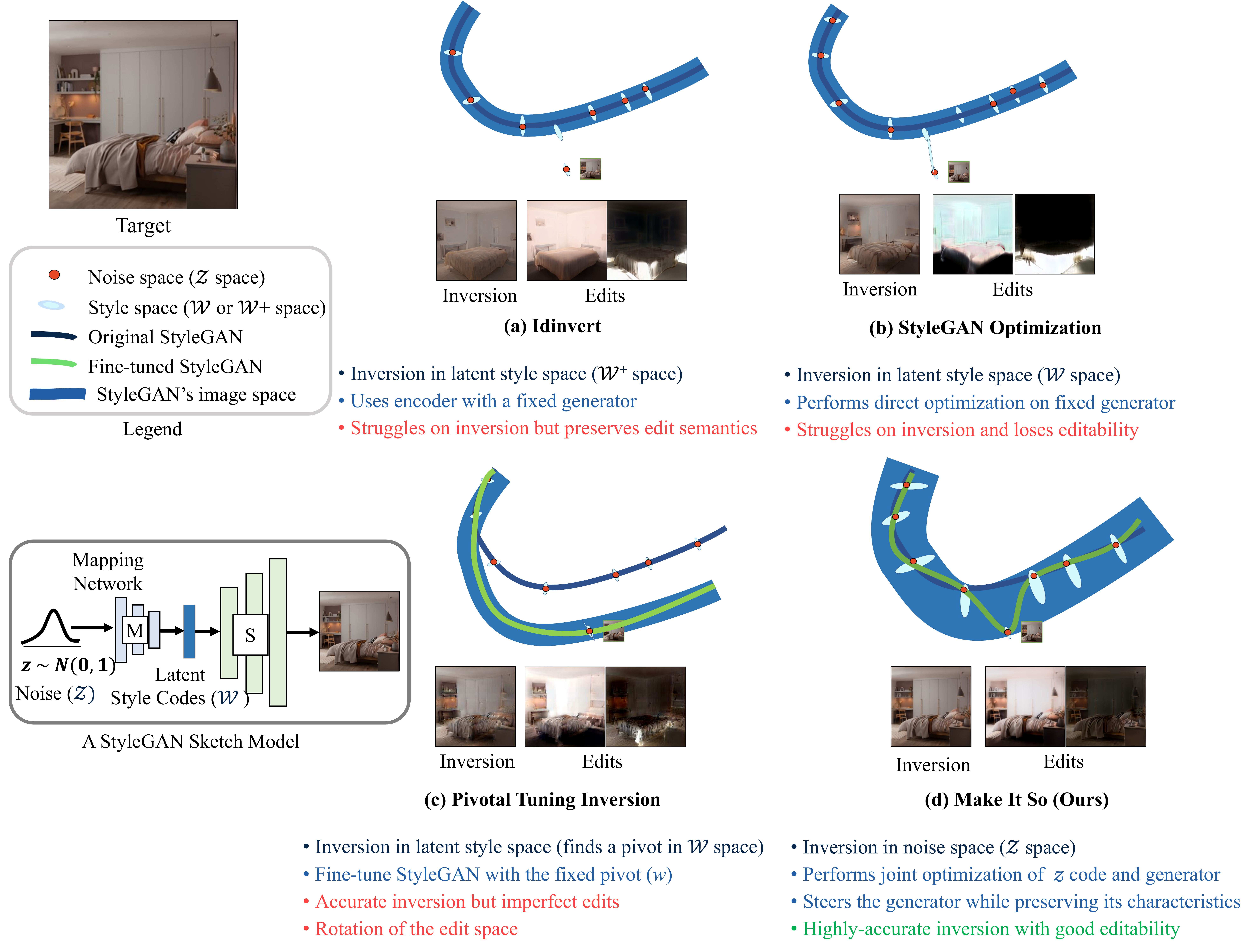}
    \vspace{-15pt}
    \caption{\textbf{A conceptual understanding of our method and distinction from prior works.} We show a 2D schematic of the noise space, latent style space, and image spaces of StyleGAN. StyleGAN produces images from noise vectors (z), shown as a 
dark blue curve.  It produces edited images from w variables manipulation (in the original
StyleGAN, w=w(z)) and these are shown as a blue region, which contains the dark
blue curve because there are strictly more such images. Individual images corresponding to a noise vector are red dots.  We hypothesize these two spaces do not cover all images (the rest of the plane), because there are images that StyleGAN is unwilling to produce.  All baseline methods invert in the latent style space ($\mathcal{W}$ or $\mathcal{W^+}$), while our method inverts in the noise space ($\mathcal{Z}$)}. 
    \label{fig:intuition}
\vspace{-18pt}
\end{figure*}

\section{Method}
Make It So combines several techniques to achieve superior image inversion and preserve editing capabilities for complex scenes. 
Firstly, the approach operates in the noise space ($\mathcal{Z}$) with joint optimization of the $z$ code and the generator, resulting in improved accuracy and editing consistency compared to prior methods that optimize the style space. Secondly, anchor and support losses are introduced as experience replay~\cite{li2017learning, de2021continual} during the optimization process to ensure that the fine-tuned StyleGAN model retains important properties of the original StyleGAN. The replay technique used is similar to those used for continual learning in a classification setting and is novel in GAN inversion.
Finally, we apply an exponential moving average strategy to decay the original StyleGAN model towards the fine-tuned StyleGAN model, enabling faster and cleaner inversion.

A summary of the conceptual differences between previous methods and Make It So is shown in Fig. \ref{fig:intuition}. Previous encoder-based inversion methods (Fig. \ref{fig:intuition}a) directly map images to StyleGAN's style space, but may not find the closest image in the image space for out-of-distribution or out-of-domain images. An alternative approach (Fig.\ref{fig:intuition}b) optimizes the style code ($w$ or $w^+$ code) to generate the image closest to the target image while keeping the StyleGAN model fixed, but may not converge to the target image and loses editing capabilities. The state-of-the-art GAN inversion approach PTI (Fig. \ref{fig:intuition}c) fine-tunes the StyleGAN generator itself by first finding the closest style code ($w$) to the target image using optimization, then fixing the style code during fine-tuning, but the inversion is not accurate, and the editing is not realistic for complex scenes. Make It So (Fig. \ref{fig:intuition}d) inverts in the noise space by jointly searching for noise and fine-tuning the StyleGAN model, achieving highly-accurate inversion and enabling various image editing tasks. The complete algorithm for Make It So is summarized in Alg. \ref{alg:make-it-so}.

\subsection{Inversion in Noise or $\mathcal{Z}$ Space}
\label{subsec:inversion-in-latent-space}
StyleGAN's style space ($\mathcal{W}$) allows for disentangled image edits. Previous methods invert images in the latent style space ($\mathcal{W}$) and risk losing editing capabilities. We propose inverting in the noise space ($\mathcal{Z}$) and finetuning the network to ensure the style space does not deviate significantly from the original model. Traditional GAN inversion techniques have explored inverting in noise space~\cite{pan2020exploiting}, but the emergence of StyleGAN has made the latent style space the go-to choice for inversion~\cite{karras2019style}. However, with appropriate losses and optimization setup, inverting in the noise space ($\mathcal{Z}$) can be advantageous for preserving the editing capabilities of the original StyleGAN model ($G_O$) while also ensuring the fine-tuned model retains desired visual attributes. This approach allows us to avoid potential loss of editing capabilities that can occur when inverting in the latent style space ($\mathcal{W}$)~\cite{roich2021pivotal}. We find that our method preserves edits for both out-of-distribution and out-of-domain images.

{\noindent \bf Joint Optimization:} Make It So fine-tunes the StyleGAN model ($G_F$) jointly with a search in noise vectors. We begin with a random noise $z$ and optimize it to minimize the reconstruction loss between the generated image $G_F(z)$ and the target image $I_t$. Concurrently, we fine-tune the StyleGAN model $G_F$ to steer it toward the target image $I_t$. 
To optimize $z$ and $G_F$ jointly, we utilize the following loss function:

\begin{equation}
\begin{aligned}
{z, G_F} = \argmin_{z, G_F} \lambda_{\text{recon}}\mathcal{L}_{\text{recon}}(G_F(z), I_t) \\ + \lambda_{\text{LPIPS}}\mathcal{L}_{\text{LPIPS}}(G_F(z), I_t)
\end{aligned}
\end{equation}
where $\lambda{\text{recon}}$ and $\lambda_{\text{LPIPS}}$ are scalar weights for the reconstruction loss and perceptual loss, respectively. The reconstruction loss is the L2 loss between the generated image $G_F(z)$ and the target image $I_t$. The perceptual loss is the LPIPS loss between the generated image $G_F(z)$ and the target image $I_t$. We use the perceptual loss to ensure that the generated image $G_F(z)$ resembles the target image $I_t$ visually. During the fine-tuning of the network, we update only the synthesis network of the StyleGAN and do not update the mapping network. This restriction ensures that the mapping network maps $z$ to the same style space $\mathcal{W}$.

\subsection{Experience Replay}
\label{subsec:learning-without-forgetting}

To maintain the important properties of the original StyleGAN model ($G_O$) required for image editing, we use experience replay~\cite{de2021continual}. This involves randomly selecting a small set of support images $I_s$ and their corresponding noise vectors $z_s$ from $G_O$ at each update iteration, as well as utilizing a small bank of edit directions $w^+_a$ from $G_O$ as an anchor. If necessary, the anchor can consist of just one edit direction.

By generating anchor and support images, we guide the fine-tuned StyleGAN model ($G_F$) towards the target image $I_t$, while still preserving the properties of the original model ($G_O$). To optimize $G_F$, we use the following losses:

\vspace{-10pt}
\begin{equation}
\begin{aligned}
G_F = \argmin_{G_F}\frac{1}{N}\sum_{i=1}^{N} \Big[\mathcal{L}_{\text{recon}}(G_F(z_s),G_O(z_s)) \\ + \mathcal{L}_{\text{LPIPS}}(G_F(z_s),G_O(z_s)) \\ + \mathcal{L}_{\text{recon}}(G_F(z_s; w^+_s+w^+_a), G_O(z_s; w^+_s+w^+a)) \\ + \mathcal{L}_{\text{LPIPS}}(G_F(z_s; w^+_s+w^+_a), G_O(z_s; w^+_s+w^+_a))\Big]
\end{aligned}
\end{equation}

Here, $w^+_s$ is the style code of the support image $I_s$ obtained from the mapping network, $N$ is the batch size, and $i$ indexes over the support images and their corresponding latent vectors in the batch. As previously mentioned, $w^+_a$ are the anchor edits. Our experiments suggest that these anchor edits can be random, but better results are achieved when using known or desired edits. We use the reconstruction loss and perceptual loss to ensure that the generated anchor and support images are visually similar to the original anchor and support images.

\subsection{Exponential Moving Average Decay}
We use an exponential moving average strategy to \emph{decay the original StyleGAN model} ($G_O$) towards the finetuned StyleGAN model ($G_F$). This enhances the stability of the inversion process and improves the quality of generated results, especially for challenging out-of-distribution or out-of-domain images.

By bringing $G_O$ closer to $G_F$, we ensure that both models possess the same properties, which is crucial since we use $G_O$ to generate support images and anchor edits for updating $G_F$. To perform the decay, we use the following equation:
\begin{equation}
\begin{aligned}
G_O = \beta G_O + (1-\beta) \times G_F
\end{aligned}
\end{equation}
Here, $\beta$ is the decay rate. We set it to $\beta=0.9999$ in our experiments. We update $G_O$ every 100 iterations of $G_F$ in our base approach (500 iterations), and every 200 iterations of $G_F$ in our extended approach (1000 iterations).

\begin{algorithm}[t]
\caption{Algorithm for Make It So. 
}
\label{alg:make-it-so}
\begin{algorithmic}[1]
\small
\Begin
\State \textbf{Input:} Target image $I_t$
\State $G_O$: Original GAN model
\State $G_F$: Fine-tuning GAN model \Comment{initialized with $G_O$}
\State $z$: noise vectors \Comment{randomly initialized}
\While{not converged}
\If{\text{EMA update:} }
\State $G_O \leftarrow \beta G_O + (1-\beta) G_F$ \Comment{decay original towards finetuned }
\EndIf
\State $z_s$: support noise vectors \Comment{random initialization}
\State $w^+_a$: editing anchors in style space \Comment{known edits}
\State $z, G_F \leftarrow \argmin{z, G_F} \mathcal{L}(G_F(z), I_t)$ \Comment{jointly optimize latent and fine-tuning model}
\State $G_F \leftarrow \argmin_{G_F} \frac{1}{N}\sum_{i=1}^{N} \Big[\mathcal{L}(G_F(z_s),G_O(z_s)) + \mathcal{L}(G_F(z_s; +w^+_a), G_O(z_s; +w^+a))\Big]$ \Comment{experience replay}
\EndWhile
\End
\end{algorithmic}
\end{algorithm}

\section{Experiments}
\setlength{\tabcolsep}{5pt}
\begin{table*}[t!]
  \caption{
    \textbf{Inversion Accuracy.} Quantitative comparisons between different inversion methods on five datasets, including LSUN Bedroom (indoor scene)~\cite{yu2015lsun}, LSUN Church (outdoor scene)~\cite{yu2015lsun}, FFHQ (human face)~\cite{stylegan}, Stanford Cars~\cite{KrauseStarkDengFei-Fei_3DRR2013}  and AFHQ Wild (animals)~\cite{choi2020stargan}. Our method recovers target images with extremely high precision and improves over previous methods by an order of magnitude. 
  }
  \label{tab:reconstruction_quat}
  \centering
  \footnotesize
  \vspace{5pt}
  \begin{tabular}{l|cc|cc|cc|cc|cc}
                \toprule
                & \multicolumn{2}{c|}{\textbf{Bedroom}}  
                & \multicolumn{2}{c|}{\textbf{Church}}  
                & \multicolumn{2}{c|}{\textbf{Face}}
                & \multicolumn{2}{c|}{\textbf{Cars}}
                & \multicolumn{2}{c}{\textbf{Animals}}
                \\
                \midrule
                & \MSE  & \LPIPS
                &    \MSE    &    \LPIPS
                &    \MSE    &    \LPIPS      
                &    \MSE    &    \LPIPS
                &    \MSE    &    \LPIPS  \\ 
                \midrule
    ALAE~\cite{pidhorskyi2020alae}
                &  0.33  &  0.65 
                &   -    &    -         
                &  0.15  &  0.32  
                & - & -
                & - & -
       \\
    IDInvert~\cite{zhu2020idinvert}
                &   0.113   &    0.41 
                &   0.140   &    0.36   
                &   0.061   &    0.22   
                & - & - 
                & - & -
         \\
    pSp~\cite{richardson2021pSp}
                &   0.099    &    0.34   
                &   0.127    &    0.31  
                &  0.034  &  0.16  
                &  0.10 & 0.29
                & 0.13 & 0.35

     \\
    e4e~\cite{tov2021e4e}
                &   -        &    -  
                &  0.142   &  0.42   
                &  0.052  &  0.20   
                &  0.12 & 0.32
                & 0.14 & 0.36

       \\
    $\text{Restyle}_{pSp}$~\cite{alaluf2021restyle}
                &   -        &    -
                &   0.090        &    0.25  
                &  0.030  &  0.13  
                & 0.07  & 0.25
                &  0.05 & 0.21
\\
    $\text{Restyle}_{e4e}$~\cite{alaluf2021restyle}
                &   -        &    -      
                &  0.129        &    0.38
                &  0.041  &  0.19   
                & 0.09  & 0.29
                & 0.07  & 0.25
 \\
    $\text{PadInv}$~\cite{Bai2022HighfidelityGI}
                &   {0.054}   &    {0.21}        
                &   {0.086}    &    {0.22} 
                & {0.021}  & {0.10}  
                & -  & - 
                & -  & -
  \\ 
    \midrule
   $\text{GHFeat}$~\cite{xu2021ghfeat}  
            & 0.068 &  - 
            & -  &  -
           & 0.046  & - 
            & - & -
            & - & -  \\
    $\text{HyperStyle}$~\cite{alaluf2022hyperstyle}
                & - & -
                & - & -
               & 0.019  & 0.09
                & 0.07 & 0.27
                & 0.06 & 0.24 
\\ 
    \midrule
    StyleGAN2~\cite{stylegan2}
                &  0.170  &   0.42   
                &  0.220  &   0.39   
                &  0.020  &   0.09 
                &  0.06 & 0.16
                & 0.03 & 0.13 
       \\
    $\text{PTI}$~\cite{roich2021pivotal} (1000 iterations)
                & 0.010 & 0.20
                & 0.012 & 0.20 
                & 0.014  & 0.09 
                & 0.01 &  0.11 
                & 0.01 & 0.08 
   \\
    Ours (500 iterations)
            &  \textbf{0.002}  &  \textbf{0.05}  
            &  \textbf{0.005}   & \textbf{0.06}  
             &  \textbf{0.002}  &   \textbf{0.02} 
            &  \textbf{0.005} & \textbf{0.09} 
            &  \textbf{0.002} & \textbf{0.07} \\
 Ours (1000 iterations)
        &  \textbf{0.002}  &  \textbf{0.03}  
        &   \textbf{0.003} & \textbf{0.03}   
        &  \textbf{0.001}  & \textbf{0.02}    
        &  \textbf{0.005} & \textbf{0.08} %
        & \textbf{0.002} & \textbf{0.05} 
      \\
                \bottomrule
  \end{tabular}
    \vspace{-10pt}
\end{table*}
In this section, we demonstrate the effectiveness of Make It So for GAN inversion on five datasets: bedrooms~\cite{yu2015lsun}, churches~\cite{yu2015lsun}, faces~\cite{stylegan}, cars~\cite{KrauseStarkDengFei-Fei_3DRR2013}, and AFHQ animals~\cite{choi2020stargan}. Successful GAN inversion requires accuracy, preserving edits, and generalization to out-of-distribution images. We provide quantitative and qualitative comparisons for inversion quality and image editing, along with various ablation studies to justify design choices such as the choice of the latent space for inversion and the choice of loss functions.

The StyleGAN models trained with the contrastive loss for bedrooms, churches, and faces datasets~\cite{yu2021dual}, and native StyleGAN models~\cite{karras2020analyzing} for Stanford Cars and AFHQ animals. Directions for our editing are obtained from ~\cite{shen2020interfacegan, shen2020sefa, wu2020stylespace, bhattad2022enriching}. Make It So is applied with a set of four support images and randomly sampled edit directions. The algorithm is run for 500 iterations with four EMA updates by default and can be applied even when only a single edit direction is desired.

Existing methods are run with their default hyperparameters and the same number of gradient updates as Make It So. For example, we use 2000 steps (900 for finding the pivot code and 1100 for fine-tuning the generator) for PTI, as opposed to 1000 iterations for Make It So, which involves both the $z$ and $G$ updates with approximately the same wall-clock time as PTI ($\approx 4$ minutes). Out-of-distribution and out-of-domain images are defined as images that are distinctly different from the training set images for a particular data domain, such as bedrooms. The majority of experiments are performed on out-of-distribution images for bedrooms to highlight the efficacy of Make It So and demonstrate generalization capabilities to other domain images from the bedroom StyleGAN model.

\subsection{Inversion Quality}
The evaluation of inversion quality is commonly conducted by computing the mean-squared error (MSE) and perceptual similarity score (LPIPS~\cite{zhang2018lpips} w/ AlexNet) between the target and inverted images. To compare inversion quality quantitatively, we consider five different datasets mentioned above. In Tab.~\ref{tab:reconstruction_quat}, we provide comprehensive quantitative comparisons with state-of-the-art GAN inversion methods, including optimization-based, encoder-based, and fine-tuning-based approaches. The MSE and LPIPS values indicate that our approach outperforms all existing methods by several orders of magnitude. Note that some of the compared approaches do not provide evaluations for all datasets that we evaluate, and therefore we report only the available numbers in Tab~.\ref{tab:reconstruction_quat}.

\begin{table}[t]\centering
\caption{{\bf Ablation and Edit Quality}. To evaluate image editing, we generate 100 random images from the original StyleGAN and apply 32 edit directions to them, creating 3200 edited images. We then invert 100 LSUN bedroom test images using both PTI and Make It So. The editing quality is evaluated by comparing the original StyleGAN images and their edits to the updated StyleGAN model after applying PTI or Make It So with the same 32 edit directions. We report the inversion quality of these test images and ablate for different components used in our method
  }
  \label{tab:ablation_losses}
\footnotesize
\vspace{5pt}
\begin{tabular}{lrrrr}\toprule
&\multicolumn{2}{c}{Inversion Quality} & \multicolumn{2}{c}{Editing Quality} \\\cmidrule{2-5}
&MSE & LPIPS &MSE &LPIPS \\\cmidrule{1-5}
PTI (1000 iterations) &0.010 &0.20 &0.360 &0.62 \\
PTI (2000 iterations) &0.008 &0.20 &0.390 &0.65 \\
\midrule
ours w/o support loss &0.003 &0.05 &\textbf{0.026} &\textbf{0.29} \\
ours w/o anchor loss &\textbf{0.002} &\textbf{0.03} &0.040 &0.36 \\
ours w/o EMA &0.003 &0.06 &0.036 &0.37 \\
ours w/o extended iterations &0.002 &0.05 &0.033 &0.34 \\
\midrule
ours  full &\textbf{0.002} &\textbf{0.03} &0.035 &0.35 \\
\bottomrule
\end{tabular}
\vspace{-15pt}
\end{table}

	\begin{figure*}
		\setlength{\linewidth}{\linewidth}
		\setlength{\hsize}{\linewidth}
		\centering
    \setlength\tabcolsep{1pt}
    \renewcommand{\arraystretch}{0.2}
    {
    \small
    \begin{tabular}{c | ccc | ccc}
		   	 \includegraphics[width=0.135\linewidth]{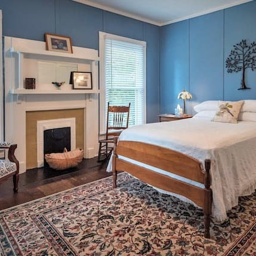}
		    & \includegraphics[width=0.135\linewidth]{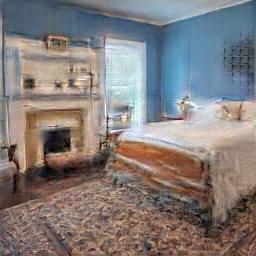}
		    & \includegraphics[width=0.135\linewidth]{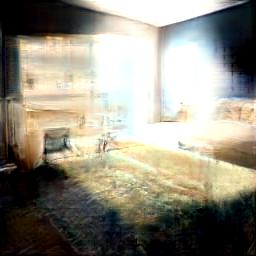}
		    & \includegraphics[width=0.135\linewidth]{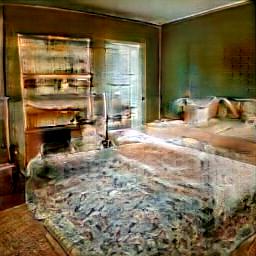}
		    &
		    \includegraphics[width=0.135\linewidth]{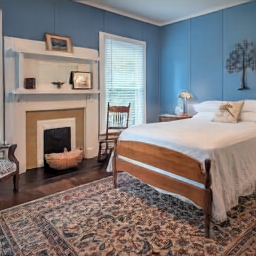}
		    &
		    \includegraphics[width=0.135\linewidth]{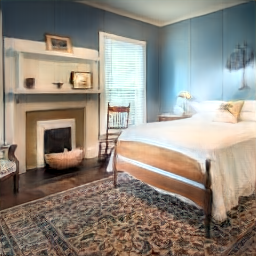}
		    & 
		    \includegraphics[width=0.135\linewidth]{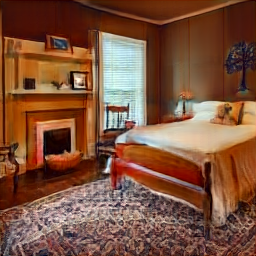}

            \\
            		    \includegraphics[width=0.135\linewidth]{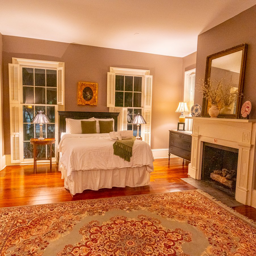}
		    &
    		    \includegraphics[width=0.135\linewidth]{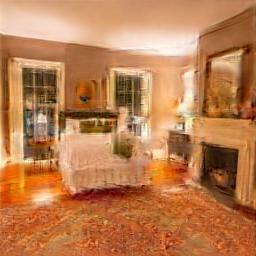}
		    &
		    \includegraphics[width=0.135\linewidth]{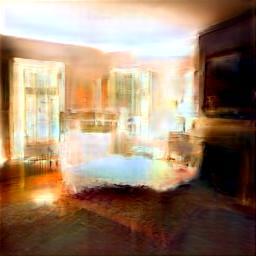}
		    & \includegraphics[width=0.135\linewidth]{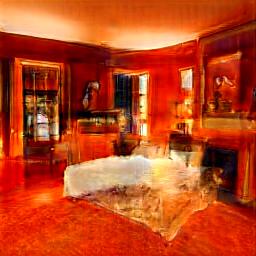}
		    &
		    \includegraphics[width=0.135\linewidth]{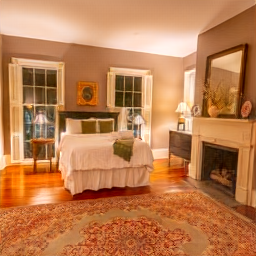}
		    &
		    \includegraphics[width=0.135\linewidth]{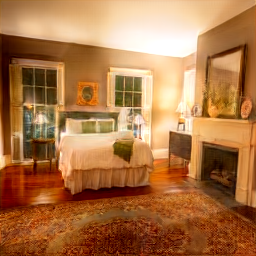}
		    &
		    \includegraphics[width=0.135\linewidth]{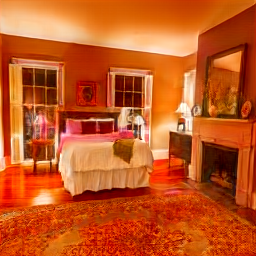}
		    \\
            \includegraphics[width=0.135\linewidth]{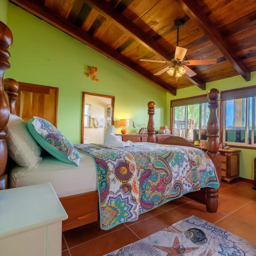}
		    & \includegraphics[width=0.135\linewidth]{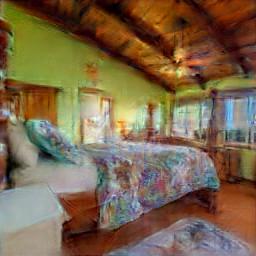}
		    & \includegraphics[width=0.135\linewidth]{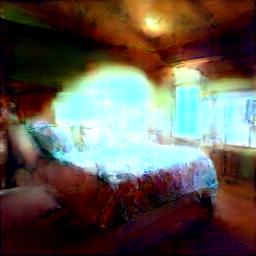}
		    & \includegraphics[width=0.135\linewidth]{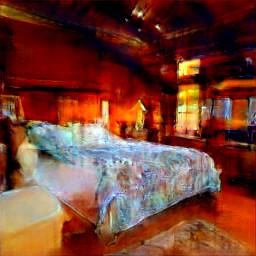}
		    &
		    \includegraphics[width=0.135\linewidth]{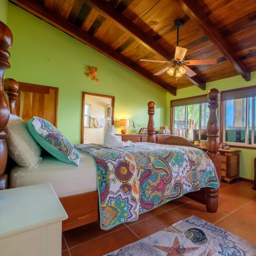}
		    &
		    \includegraphics[width=0.135\linewidth]{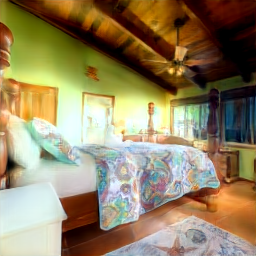}
		    &
		     \includegraphics[width=0.135\linewidth]{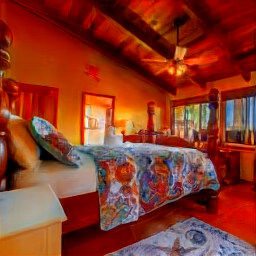}

            \vspace{2pt}

		    \\

            \scriptsize{(1) Target Scene}& 
            \scriptsize{(2) Inversion}&
            \scriptsize{(3) Relighted}&
            \scriptsize{(4) Resurfacing}&
            \scriptsize{(5) Inversion}&
            \scriptsize{(6) Relighted}&
            \scriptsize{(7) Resurfacing}
            \vspace{5pt}
            \\
            &  \multicolumn{3}{c|}{Pivotal Tuning Inversion (PTI)~\cite{roich2021pivotal}} & \multicolumn{3}{c|}{Make It So (ours)}
            \vspace{2pt}
        \end{tabular}
\caption{{\bf Qualitative comparison}. We compare with the current SotA method, PTI, that also tunes the generator. 
Column 1 shows the target scene. Columns 2--4 show the results obtained by using PTI. Columns 5--7 show the results obtained by using Make It So. {\bf Inversion quality:} with Make it So high-precision inversion is achieved with no visible artifacts that are evident in PTI (Column 2).  {\bf Editing quality:} it is evident that PTI has difficulty preserving desired edits, and the images appear unrealistic. Make It So operates on remembering good properties of StyleGAN and ensuring that the $\mathcal{W}$ space is preserved and transferred to the finetuned model. As a result, the images can be edited realistically without loss of generality.
}
\label{fig:PTIComparison}
}
\vspace{-3pt}
\end{figure*}

\begin{figure*}[t]
		\setlength{\linewidth}{\linewidth}
		\setlength{\hsize}{\linewidth}
		\centering
    \setlength\tabcolsep{3pt}
    \renewcommand{\arraystretch}{0.2}
    \begin{tabular}{c|cc|cc|cc}
\includegraphics[width=0.13\linewidth]{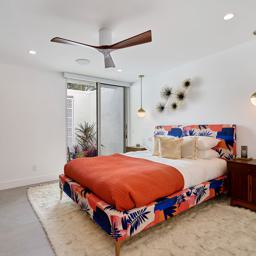} & 
\includegraphics[width=0.13\linewidth]{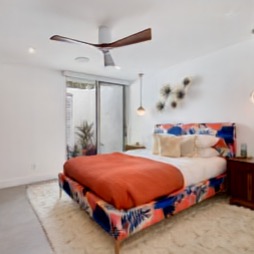} & 
\includegraphics[width=0.13\linewidth]{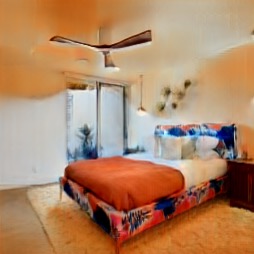} &
\includegraphics[width=0.13\linewidth]{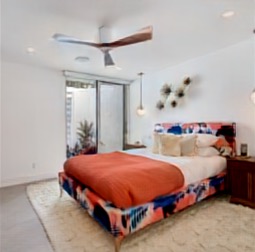} &
\includegraphics[width=0.13\linewidth]{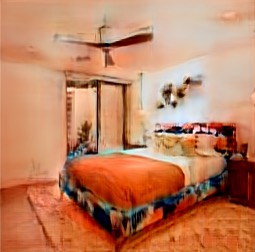}&
\includegraphics[width=0.13\linewidth]{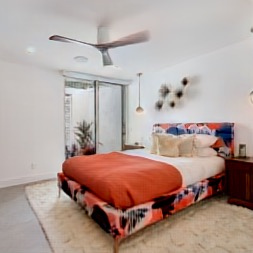} &
\includegraphics[width=0.13\linewidth]{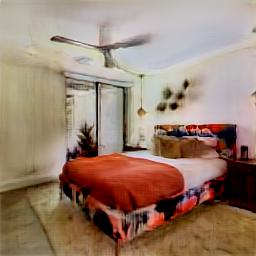}
 
\\
\includegraphics[width=0.13\linewidth]{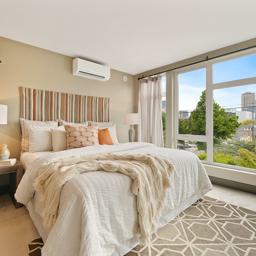} & 
\includegraphics[width=0.13\linewidth]{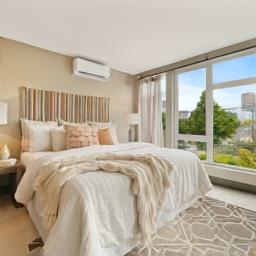} & 
\includegraphics[width=0.13\linewidth]{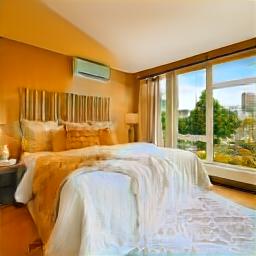} &
\includegraphics[width=0.13\linewidth]{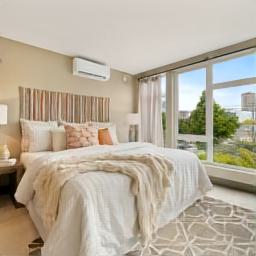} &
\includegraphics[width=0.13\linewidth]{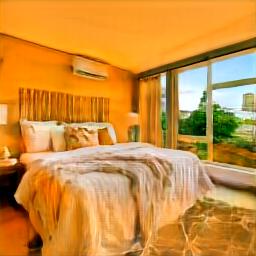}&
\includegraphics[width=0.13\linewidth]{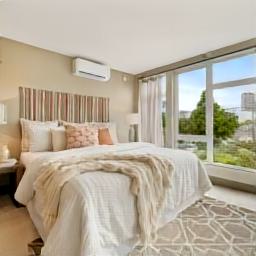} &
\includegraphics[width=0.13\linewidth]{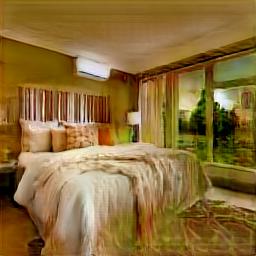} 
            \vspace{2pt}

\\

            \scriptsize{Target Scene}& 
            \multicolumn{2}{c}{\scriptsize{Inversion (left) \& editing (right) in $\mathcal{Z}$ space}}
            &
             \multicolumn{2}{c}{\scriptsize{Inversion (left) \& editing (right) in $\mathcal{W}$ space}}
          &
            \multicolumn{2}{c}{\scriptsize{Inversion (left) \& editing (right) in $\mathcal{W}+$ space}}
            \vspace{3pt}
        \end{tabular}
\caption{{\bf Ablation on Choices of Inversion Space}. We compare Make It So inversions when operating in different spaces such as $\mathcal{Z, W}$ and $\mathcal{W^+}$ space. The result clearly shows that inverting in $\mathcal{Z}$ space provides the best quality results for both inversion and editing. Inverting in $\mathcal{W}$ and $\mathcal{W^+}$ spaces adversely affects the editing capabilities of the generator and performs worse for editing.}
\label{fig:latent_space_choice}
\vspace{-12pt}
\end{figure*}

\subsection{Quality of Editing}

Producing accurate inversion by overfitting the target image is a trivial task for a tunable generator. Therefore, it is crucial to evaluate the usefulness of inversion for image editing. Our approach ensures that the editing capabilities of the generator are preserved during fine-tuning, enabling us to use pre-calculated editing directions on the fine-tuned generator for performing image editing. In Fig.\ref{fig:teaser}, In Fig.\ref{fig:bedroom_teaser} and Fig.\ref{fig:PTIComparison}, we present results on various edits on out-of-distribution bedroom images. It is evident that our approach produces better edits compared to PTI. Particularly, for global edits such as relighting and resurfacing, PTI lacks the semantic understanding of various objects in the image, while Make It So maintains consistent semantics in its edits. 

 In Fig.~\ref{fig:PTIComparison} and Fig.~\ref{fig:bedroom_teaser}, we present results on various edits on out-of-distribution bedroom images. It is evident that our approach produces better edits compared to PTI. Particularly, for global edits such as relighting and resurfacing, PTI lacks the semantic understanding of various objects in the image, while Make It So maintains consistent semantics in its edits.

\begin{figure*}[t]
		\setlength{\linewidth}{\linewidth}
		\setlength{\hsize}{\linewidth}
		\centering
    \setlength\tabcolsep{1pt}
    \renewcommand{\arraystretch}{0.2}
    \begin{tabular}{ccccccc}
\includegraphics[width=0.135\linewidth]{figures/bedroom/electric/orig.png} & 
\includegraphics[width=0.135\linewidth]{figures/bedroom/electric/makeitso.png} & 
\includegraphics[width=0.135\linewidth]{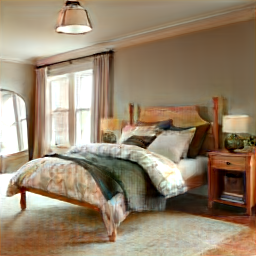} & 
\includegraphics[width=0.135\linewidth]{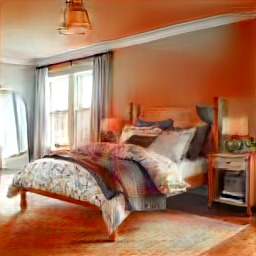} &
\includegraphics[width=0.135\linewidth]{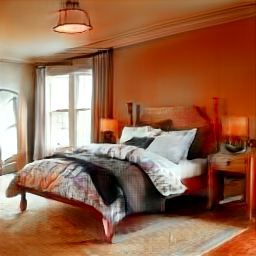} &
\includegraphics[width=0.135\linewidth]{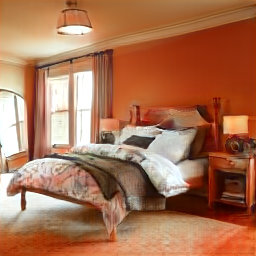} &
\includegraphics[width=0.135\linewidth]{figures/bedroom/electric/electric_orange.png}
            \vspace{2pt}
\\

            \scriptsize{Target Scene}& 
             \scriptsize{Inversion}& 
            \scriptsize{Finetune Network}&
            \scriptsize{+Support Loss}&
            \scriptsize{+Anchor Loss}&
            \scriptsize{+EMA Decay} &
            \scriptsize{+Extended Iterations} 
\vspace{5pt}
        \end{tabular}
\caption{{\bf Ablation for Editing}. For images in the first column, we first show Make It So inversion in the second column. We then ablate for each component used in Make It So incrementally for their ability to preserve edits starting from the third column. We add a new component to the previous column's setup from left to right. With finetuning network alone, we recover scene layout but not edits. With our support loss, in combination with finetuning, we observe moderate edits. The  addition of anchor loss results in strong edits, but overall we still cannot achieve high-quality inversion. A periodic decay of the original StyleGAN model towards the finetuned model improves our inversion quality by preserving edits. The use of an extended iteration period (from 500 to 1000 iterations) allows us to recover finer details with greater precision. 
The results improve from left to right as fine details are preserved. Observe patches around table lamps.}
\label{fig:contribution_ablations}
	\end{figure*}

\begin{figure*}[t!]
	\scriptsize
		\centering
    \setlength\tabcolsep{0.5pt}
    \renewcommand{\arraystretch}{0.2}
    \begin{tabular}{c|cc|cc|cc|cc|cc}
\includegraphics[width=0.085\linewidth]{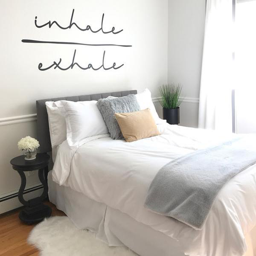} & 
\includegraphics[width=0.085\linewidth]{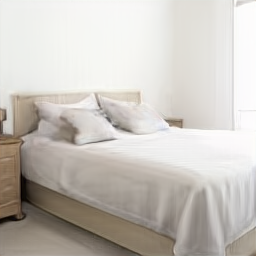} & 
\includegraphics[width=0.085\linewidth]{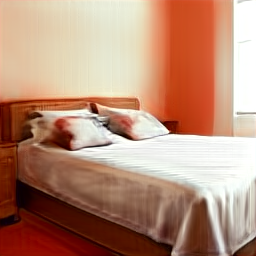} &
\includegraphics[width=0.085\linewidth]{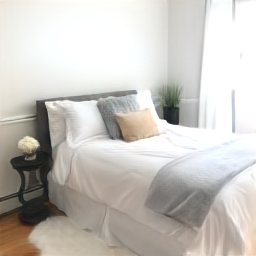} & 
\includegraphics[width=0.085\linewidth]{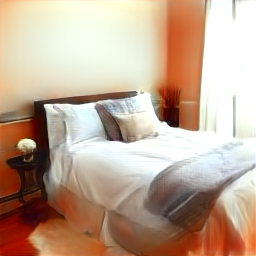} &
\includegraphics[width=0.085\linewidth]{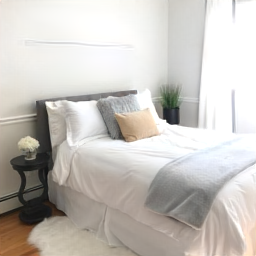} & 
\includegraphics[width=0.085\linewidth]{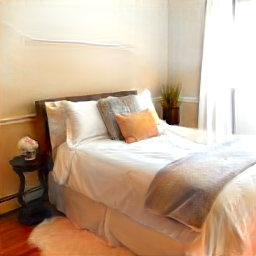} &
\includegraphics[width=0.085\linewidth]{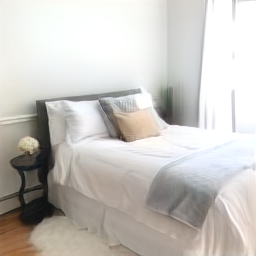} &
\includegraphics[width=0.085\linewidth]{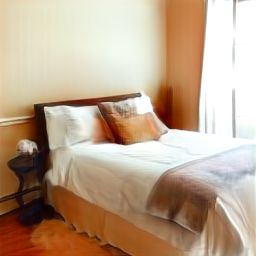} &
\includegraphics[width=0.085\linewidth]{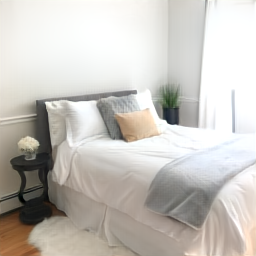} & 
\includegraphics[width=0.085\linewidth]{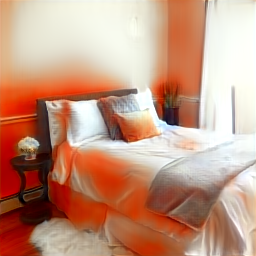} 
\\

            \scriptsize{Target Scene}& 
             \multicolumn{2}{c|}{\scriptsize{Inversion \& Edited}}
            &
             \multicolumn{2}{c|}{\scriptsize{Inversion \& Edited}}
            &              \multicolumn{2}{c|}{\scriptsize{Inversion \& Edited}}
            &              \multicolumn{2}{c|}{\scriptsize{Inversion \& Edited}}
            &              \multicolumn{2}{c}{\scriptsize{Inversion \& Edited}}\\
            & \multicolumn{2}{c|}{\scriptsize{w/o finetune network}} 
             & \multicolumn{2}{c|}{\scriptsize{w/o support loss}} 
              & \multicolumn{2}{c|}{\scriptsize{w/o anchor loss}} 
                & \multicolumn{2}{c|}{\scriptsize{w/o extended iterations}}
                 & 
                  \multicolumn{2}{c}{\scriptsize{full}} 
\vspace{3pt}
        \end{tabular}
\caption{{\bf Leave-One-Out Ablation}. We demonstrate a challenging leave-one-out ablation analysis. Good edits are obtained when using no support loss, but the inversion is not near-perfect. Our full pipeline produces a better inversion, but the edit is slightly worse compared to results without support loss, which is consistent with our Tab. 2. This instance can also be considered a failure case when 1000 iterations are insufficient for near-perfect inversion.}
\label{fig:leave_one_ablation}
\vspace{-12pt}
\end{figure*}

\noindent{\bf Preserving the Edit Space.} Fig.~\ref{fig:contribution_ablations} shows how each individual component, including the losses, exponential decay strategy, extended iterations, and use of multiple hooks, helps preserve the edit space while ensuring faster and cleaner inversion. Additionally, we provide a leave-one-out ablation in Fig. \ref{fig:leave_one_ablation}. In Tab.~\ref{tab:ablation_losses}, our results demonstrate that Make It So outperforms PTI in preserving edit quality by a factor of 10. We emphasize that our anchor loss is critical for maintaining the edit space's integrity and using it as is on the inverted noise vectors ($z$) and the updated GAN model ($G_F$). 

We also show ablation for the choice of Inversion space that should be adopted for inversion in Fig.~\ref{fig:latent_space_choice}. It supports our intuition for inverting in noise space $\mathcal{Z}$. There is visible interaction between the edit codes and inverted codes when inverting in latent style space ($\mathcal{W}$ or $\mathcal{W}^+$).

\begin{figure}[htpb!]
		\setlength{\linewidth}{\linewidth}
		\setlength{\hsize}{\linewidth}
		\centering
    \setlength\tabcolsep{0.5pt}
    \renewcommand{\arraystretch}{0.2}
    {
    \small
    \begin{tabular}{c | ccc | ccc}
		    \includegraphics[width=0.135\linewidth]{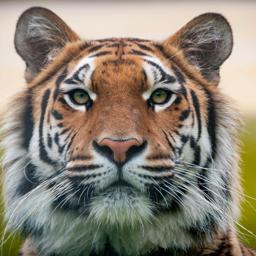}
		    &
		     \includegraphics[width=0.135\linewidth]{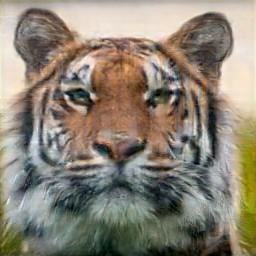}
		    &
		     \includegraphics[width=0.135\linewidth]{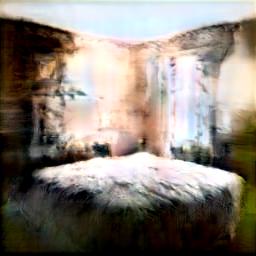}
		    &
             \includegraphics[width=0.135\linewidth]{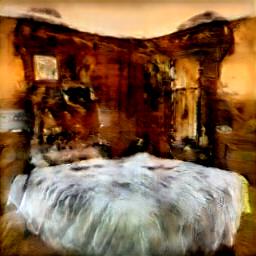}
		    &
		    \includegraphics[width=0.135\linewidth]{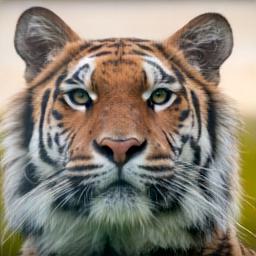}
		    &
		    \includegraphics[width=0.135\linewidth]{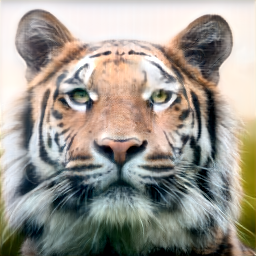}
		    &
		    \includegraphics[width=0.135\linewidth]{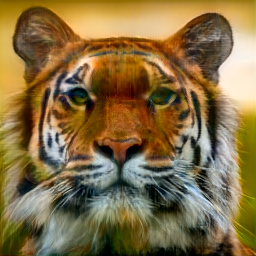}
		     \\
		   	 \includegraphics[width=0.135\linewidth]{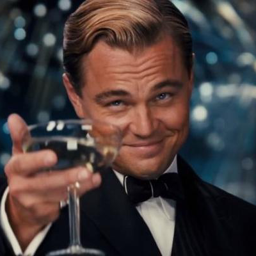}
		    & \includegraphics[width=0.135\linewidth]{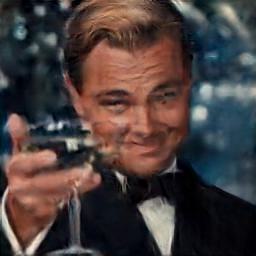}
		    & \includegraphics[width=0.135\linewidth]{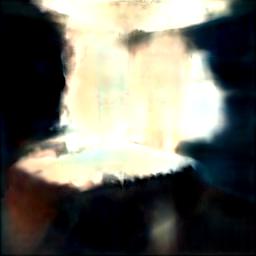}
		    & \includegraphics[width=0.135\linewidth]{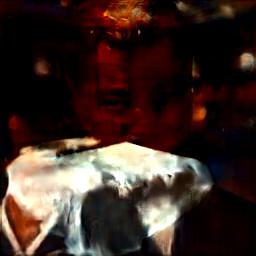}
		    &
		    \includegraphics[width=0.135\linewidth]{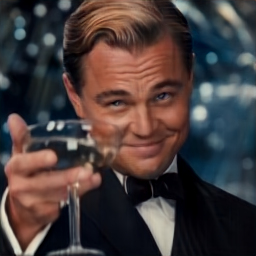}
		    &
		    \includegraphics[width=0.135\linewidth]{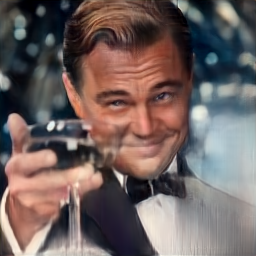}
		    & 
		    \includegraphics[width=0.135\linewidth]{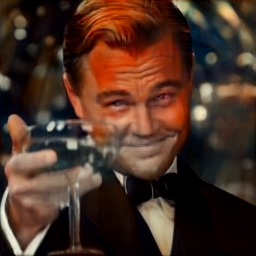}

            \\
		   	 \includegraphics[width=0.135\linewidth]{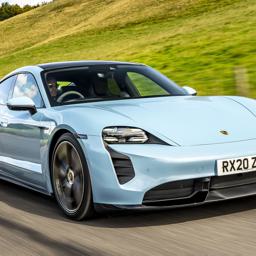}
		    & \includegraphics[width=0.135\linewidth]{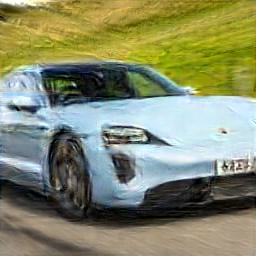}
		    & \includegraphics[width=0.135\linewidth]{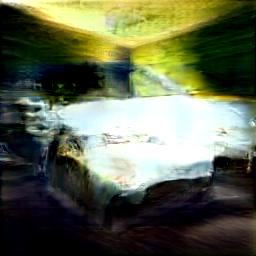}
		    & \includegraphics[width=0.135\linewidth]{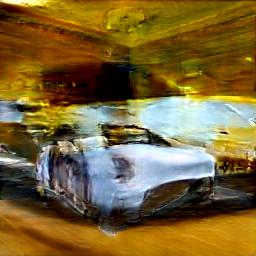}
		    &
		    \includegraphics[width=0.135\linewidth]{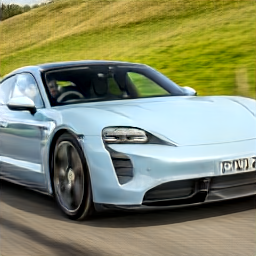}
		    &
		    \includegraphics[width=0.135\linewidth]{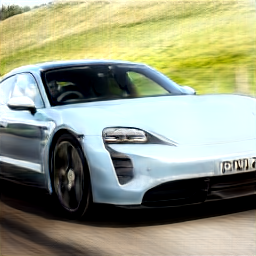}
		    & 
		    \includegraphics[width=0.135\linewidth]{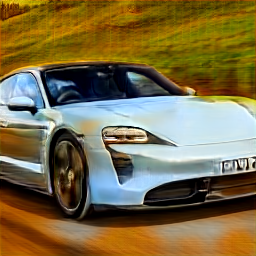}

            \\

		    \\

            \scriptsize{Target}& 
            \scriptsize{Inverted}&
            \scriptsize{Relighted}&
            \scriptsize{Recolored}&
            \scriptsize{Inverted}&
            \scriptsize{Relighted}&
            \scriptsize{Recolored}
            \vspace{5pt}
            \\
            &  \multicolumn{3}{c|}{\scriptsize{PTI}} & \multicolumn{3}{c|}{\scriptsize{Make It So (ours)}}
            \vspace{5pt}
        \end{tabular}
\caption{{\bf Additional Out-of-Domain Generalization Results}. Make It So utilizes a single StyleGAN model trained exclusively on bedroom images to invert a wide range of out-of-domain and out-of-distribution images, including human faces, cars, and animals, without requiring additional training. The resulting images exhibit realistic inversion as well as reasonable global edits, such as relighting complex scenes. In contrast, PTI generates unrealistic edits by attempting to transform out-of-domain images into bedrooms, as evidenced by the wall-like patterns in the background and bed-like patterns in the foreground of PTI edits.}
\vspace{-12pt}
\label{fig:rebood1}
}

\end{figure}

\subsection{Generalization to Out-of-Domain Images}
\label{subsec:generalization}

The current approach for image inversion and editing with StyleGAN involves using category-specific models, which makes extending StyleGAN's editing capacity to out-of-domain datasets a complex task. However, Make It So exploits the fact that StyleGAN serves as an image-prior and encodes semantic properties within its learned representation, enabling editing. We aim to invert out-of-domain images using a single StyleGAN model by transferring its edit space. To test this, we use a bedroom StyleGAN model that models complex spatial relationships with different geometry and materials. We experiment with several out-of-domain examples, including churches, capitol buildings, cars with complex material properties, animals with fine spatial details, and human faces. Figures \ref{fig:out-of-domain-generalization} and \ref{fig:rebood1} display the clean inversion for these out-of-domain images, demonstrating the capability of our approach. Moreover, we show that we can apply edits used for bedrooms to other images realistically.

We also experimented with using the face StyleGAN model as the base model for inverting and editing out-of-domain images of rooms, animals, and cars. However, we found this approach to be ineffective. In contrast, using a room StyleGAN model as the base for inverting and editing out-of-domain images of faces, animals, and cars produced favorable results. We hypothesize that this is because rooms are ``visually richer'' than faces, although it is challenging to predict visual richness with certainty. In our future work, we plan to explore the generalizability of other StyleGAN models to further investigate this hypothesis.

\section{Conclusion}
In conclusion, we introduced Make It So, a novel approach for image inversion and editing using StyleGAN. Make It So's novelty is not based on a single new technique, but rather a combination of methods that result in superior inversion and preservation of editing capabilities. Furthermore, Make It So can generalize and successfully invert and edit out-of-domain images, including Capitol buildings, cars, animals, and human faces, using a single GAN model, a property not observed in previous state-of-the-art methods. Our ablation study highlighted the importance of each individual component of our approach in preserving the edit space while ensuring clean and faster inversion. We concluded that inverting in the noise space ($\mathcal{Z}$) is the best choice for image inversion. However, a major limitation of Make It So, like PTI, is that it is not a real-time approach due to its optimization-based nature. Additionally, for extremely challenging scenes, it may be necessary to use more updates to achieve near-perfect inversion (Fig.~\ref{fig:leave_one_ablation}). In summary, Make It So offers an effective solution to the challenging problem of image inversion and editing, providing significant improvements for complex scenes. Future work will explore the generalizability of other StyleGAN models.

{\small
\bibliographystyle{ieee_fullname}
\bibliography{virajbib,egbib}
}
\clearpage
\end{document}